\documentclass[10pt,twocolumn,letterpaper]{article}

\usepackage{titling}
\usepackage[pagenumbers]{cvpr} 

\usepackage{graphicx}
\usepackage{amsmath}
\usepackage{amssymb}
\usepackage{booktabs}

\usepackage{url}            
\usepackage{amsfonts}       
\usepackage{nicefrac}       
\usepackage{microtype}      
\usepackage{xcolor}         
\usepackage{adjustbox}
\usepackage{longtable}
\usepackage{lscape}
\usepackage{placeins}
\usepackage{multirow}
\usepackage{multicol}
\usepackage{mathtools}
\usepackage{algorithm}
\usepackage{algorithmic}
\usepackage{rotating}

\newcommand{\myparagraph}[1]{\vspace{1pt}\noindent{\bf{#1}}}

\newcommand{\fullnameBold}{\textbf{A}ttention \textbf{C}onsistency on \textbf{V}isual \textbf{C}orruptions}
\newcommand{\fullname}{Attention Consistency on Visual Corruptions}
\newcommand{\nickname}{ACVC}

\newcommand{\nicknameCorruptions}{VC}

\usepackage[pagebackref,breaklinks,colorlinks]{hyperref}

\usepackage[capitalize]{cleveref}
\crefname{section}{Sec.}{Secs.}
\Crefname{section}{Section}{Sections}
\Crefname{table}{Table}{Tables}
\crefname{table}{Tab.}{Tabs.}

\usepackage{cleveref}

\def\cvprPaperID{61} 
\def\confName{CVPR}
\def\confYear{2022}

\begin{document}

\title{Attention Consistency on Visual Corruptions\\for Single-Source Domain Generalization}

\author{Ilke Cugu$^1$, Massimiliano Mancini$^1$, Yanbei Chen$^1$, Zeynep Akata$^{1,2}$\\
$^1$University of T\"{u}bingen, $^2$MPI for Intelligent Systems\\
{\tt\small \{ilke.cugu, massimiliano.mancini, yanbei.chen, zeynep.akata\}@uni-tuebingen.de}
}
\maketitle

\begin{abstract}
Generalizing visual recognition models trained on a single distribution to unseen input distributions (\ie domains) requires making them robust to superfluous correlations in the training set. In this work, we achieve this goal by altering the training images to simulate new domains and imposing consistent visual attention across the different views of the same sample. We discover that the first objective can be simply and effectively met through visual \textit{corruptions}. Specifically, we alter the content of the training images using the nineteen corruptions of the ImageNet-C benchmark and three additional transformations based on Fourier transform. Since these corruptions preserve object locations, we propose an attention consistency loss to ensure that class activation maps across original and corrupted versions of the same training sample are aligned. We name our model \fullnameBold\ (\nickname). We show that \nickname\ consistently achieves the state of the art on three single-source domain generalization benchmarks, PACS, COCO, and the large-scale DomainNet \footnote{The codes are available at \url{https://github.com/ExplainableML/ACVC}}.
\end{abstract}

\section{Introduction}
\label{sec:intro}

Visual recognition models aim to categorize the semantic content of an image. While existing deep learning methods have achieved impressive results on standard object recognition benchmarks \cite{krizhevsky2012imagenet,he2016deep}, their performance degrades when the test data distribution differs from the training one \cite{wang2018deep}. This problem, called \textit{domain-shift} \cite{csurka2017comprehensive} is ubiquitous for systems operating in real environments. In fact, since we cannot collect data for every possible change in the input distribution (\eg illumination, background, weather, etc.), we need to develop models that can generalize to \textit{unseen} domains (\ie input distribution) not represented in the training set.

Towards this goal, in this paper, we address the problem of \textit{single-source domain generalization} (single DG), where only a single (source) domain is available for training, and multiple unseen domains are present at test time. This problem is challenging since, contrary to standard domain generalization, we cannot rely on multiple training domains to disentangle domain-specific and domain-invariant information \cite{li2017deeper,motiian2017unified,chattopadhyay2020learning,wang2020learning}. This led previous approaches to simulate multiple domains via data augmentation and adversarial perturbations \cite{volpi2018generalizing,qiao2020learning,zhaoNIPS20maximum}, using them within standard classification objectives \cite{volpi2018generalizing,volpi2019addressing,zhaoNIPS20maximum}, or meta-learning procedures \cite{qiao2020learning}.

\begin{figure}
\centering
\includegraphics[scale=0.48]{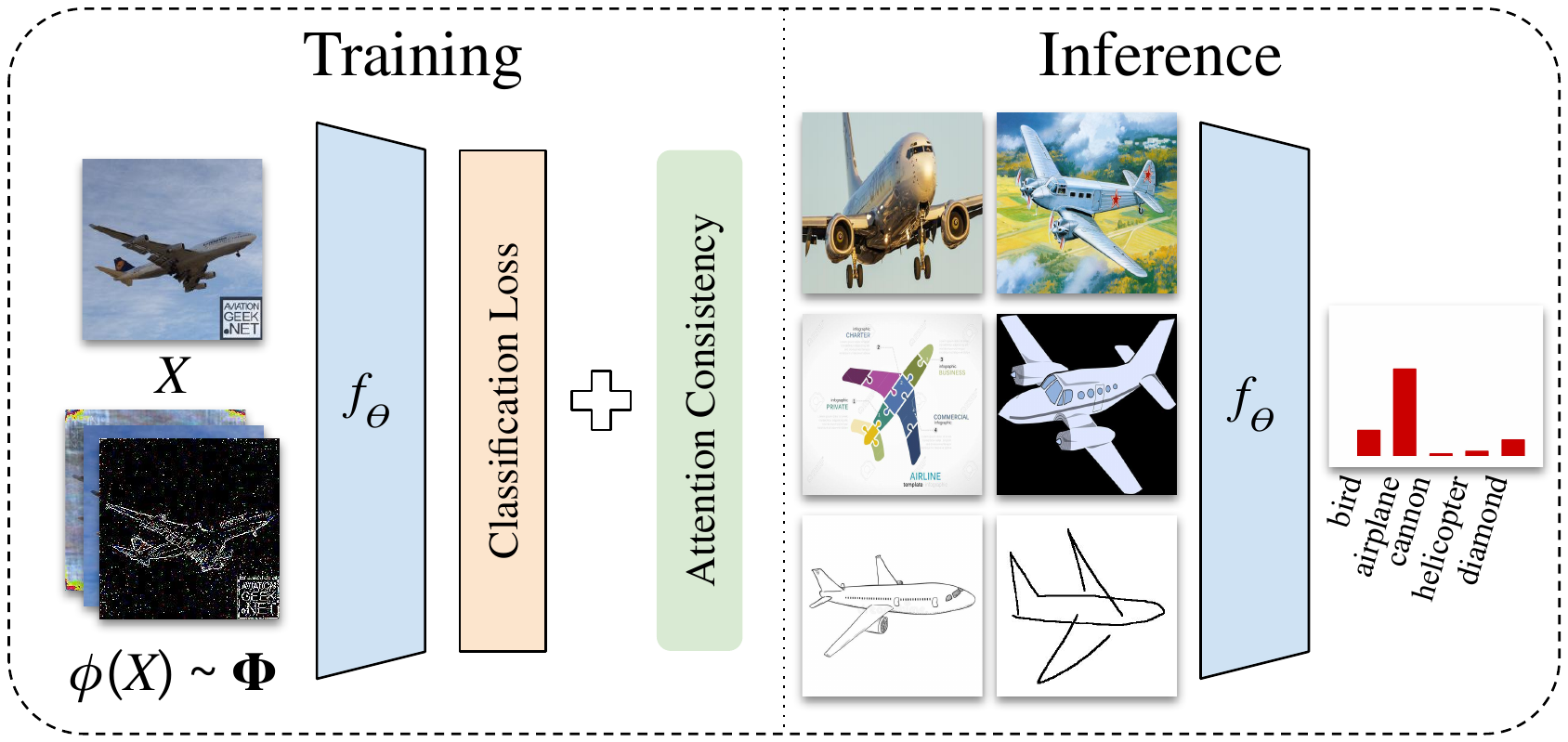}
\caption{Our approach (1) samples a transformation from a pool of visual corruptions (i.e. $\phi(X) \sim \Phi$) to simulate distinct domains for training, and (2) enforces visual attention consistency between the original and corrupted sample. Once trained, our model is capable of generalizing well to unseen domains.}
\label{fig:VisCo_teaser}
\end{figure}

Training with multiple synthetic domains allows the model to better disentangle domain- and semantic-specific information, eliminating spurious correlations between the model's predictions and the input images. Here we start from the same principle, \ie augmenting data to simulate different training domains. However, we take a step further and we argue that a robust single DG model should provide the same explanation across augmented views of the same training sample. In particular, we compute the model's Class Activation Maps \cite{zhou2016learning} for both the original and augmented samples, imposing consistency among the two (Figure \ref{fig:VisCo_teaser}). This forces the model to look at the same spatial locations, no matter how different the augmented sample looks like. We found this approach to provide a stronger learning signal in comparison to alignment on model predictions \cite{hendrycks2020augmix}. 

\begin{figure*}
\centering
\includegraphics[scale=0.38]{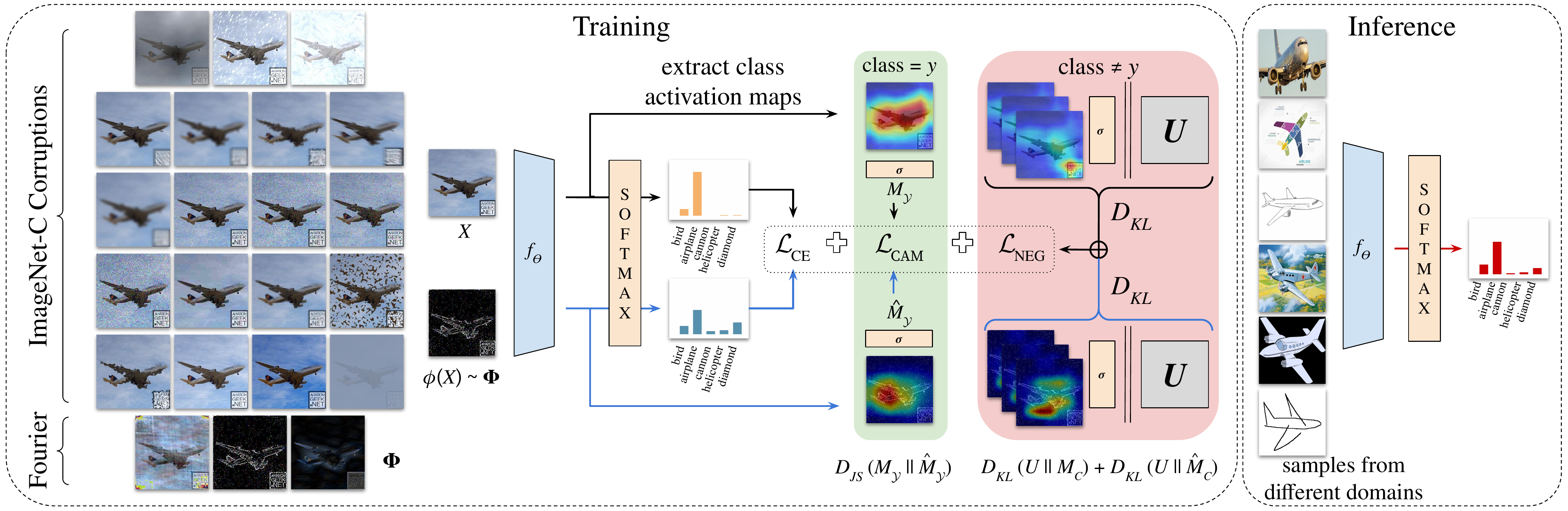}
\caption{Our \nickname\ approach (1) randomly samples a corruption $\phi$ from the set of twenty-two augmentations $\Phi$ that consist of ImageNet-C and our Fourier-based corruptions, (2) enforces visual attention consistency between a given model's class activation maps (CAM) for the original $M_y$ and corrupted version $\hat{M_y}$ of a given image $X$, (3) regularize the CAMs via Negative CAM loss \cite{sun2020fixing} that minimizes the difference between uniform distribution $U$ and top-$k$ negative classes' CAMs $M_{c \in C_k}$.}
\label{fig:VisCo}
\end{figure*}

Another crucial element of our framework is the data augmentation technique. It should heavily alter the input while not modifying the spatial location of the semantic content. To achieve this, we propose to use visual corruptions. Our idea is that corrupting the images not only creates different input domains, but also produces abundant task-irrelevant visual variations, which together help to prevent the model from memorizing spurious patterns in the training set. We make use of five families of visual corruptions (shown in Figure~\ref{fig:VisCo}), \ie Weather, Blur, Noise, Digital, and Fourier. The first four groups contain transformations taken from the ImageNet-C \cite{hendrycks2018benchmarking} benchmark. The last group contains three transformations corrupting the image using the post-Fourier transform components (Figure~\ref{fig:VisCo}, bottom left) by removing low frequencies, modifying amplitudes, and scaling phases. 

To summarize, our contributions are as follows. (1) We analyze the use of visual corruptions as augmentation technique for single DG, using $19$ transformations drawn from ImageNet-C and $3$ Fourier-based ones. (2) We propose a new consistency loss based on class activation maps, forcing the model to look at the same regions for both the clean and corrupted images (Figure \ref{fig:VisCo}, green box)). We name our model \fullnameBold\ (\nickname). (3) We propose a new single DG benchmark using three different datasets: PACS \cite{li2017deeper}, COCO \cite{lin2014microsoft} and DomainNet \cite{peng2019moment}, that measure generalization performance of models from natural images to other domains; (4) We show that \nickname\ achieves the state-of-the-art on the proposed single-source DG benchmarks, outperforming information-bottleneck based adversarial (\eg ME-ADA \cite{zhaoNIPS20maximum}) and advanced data augmentation techniques (\eg MixUp \cite{zhang2017mixup}, CutMix \cite{yun2019cutmix}, CutOut \cite{devries2017improved}, RandAugment \cite{cubuk2020randaugment}, and AugMix \cite{hendrycks2020augmix}). 

\section{Related Work}
\label{sec:related_works}

\myparagraph{Domain generalization} (DG) is the task of learning a model that generalizes to data distributions unseen during training \cite{ghifary2015domain,li2017deeper}. While this problem is usually addressed in the multi-source setting, here we focus on the scenario where only a single domain is available during training \cite{qiao2020learning}, \ie single DG. This is challenging since we cannot rely on the presence of multiple training domains to \eg disentangle domain-specific and domain-invariant information \cite{li2017deeper,chattopadhyay2020learning,seo2020learning,chen2021style}, or align feature distributions of different domains while preserving their semantics \cite{li2018domain,motiian2017unified,carlucci2019domain,zhou2020domain}. Typical approaches for single DG simulate the presence of new domains with data augmentation either through adversarial strategies \cite{li2021progressive,volpi2018generalizing,qiao2020learning,zhaoNIPS20maximum,qiao2021uncertainty,fan2021adversarially} or direct input transformation \cite{volpi2019addressing}. For instance, \cite{volpi2018generalizing} performs adversarial data augmentation under a worst case formulation, assuming samples of unseen domains to be close to the training distribution. \cite{qiao2020learning} relaxes the worst-case formulation of \cite{volpi2018generalizing} through Wasserstein Auto-Encoders \cite{tolstikhin2018wasserstein}, using the augmented domains to perform meta learning. \cite{zhaoNIPS20maximum} uses information bottleneck (IB) principle \cite{tishby2000information} to generate adversarial samples far from the source domain. \cite{volpi2019addressing} defines new data augmentation rules through an evolutionary strategy, with the fitness measure being the model error.

Differently from these works, we focus on \textit{corruptions} of the input images as transformations. We show that removing information from the data provides better generalization performance than more complex data augmentation schemes. Moreover, we are the first to use visual explanation techniques as consistency loss for DG, enforcing the model to attend to the same regions, regardless the style of the input.

\myparagraph{Data augmentation} is an effective strategy to improve the generalization of deep neural networks, providing different views of the same input. In computer vision, the most common augmentation strategies are label-preserving transformations such as random flipping, cropping and rotations~\cite{krizhevsky2012imagenet,ciregan2012multi}. Recently, various advanced augmentation techniques have been proposed to further improve representation learning, including CutOut \cite{devries2017improved}, CutMix \cite{yun2019cutmix}, MixUp \cite{zhang2017mixup} and automated augmentation schemes such as AutoAugment \cite{cubuk2018autoaugment}, RandAugment \cite{cubuk2020randaugment}, and AugMix \cite{hendrycks2020augmix}. In these techniques, an input image is often randomly corrupted by mixing with another image (\eg CutMix \cite{yun2019cutmix}, MixUp \cite{zhang2017mixup}) or by random occlusion (\eg CutOut \cite{devries2017improved}). Such corruptions, however, may destroy the underlying semantics of the input image and even alter its corresponding class label \cite{zhang2017mixup,yun2019cutmix}. In automated augmentation, augmentation strategies are either learned w.r.t. the performance on the validation set \cite{cubuk2018autoaugment}, or randomly selected from a pool \cite{cubuk2020randaugment,hendrycks2020augmix}. In addition, AugMix \cite{hendrycks2020augmix} uses a Jensen-Shannon divergence loss on model's predictions for the original and augmented images. 

In this work, we propose to use a diverse set of visual corruptions randomly selected per image during training. Since our transformations alter neither the semantic of the image nor the location of the objects, we formulate a visual attention consistency loss to encourage the model to look at the same regions for both the original and corrupted versions of a given image.

\section{\fullname}
\label{sec:method}

We aim to solve the problem of single domain generalization (single DG) where a 
model is trained on data from a single domain (source) but is expected to generalize to domains unseen during training (target). Formally, we are given a training set $\mathcal{D}=\{(x_i,y_i)\}_{i=1}^N$, where $x\in\mathcal{X}$ is an image in the space $\mathcal{X}$ and $y$ is its corresponding class label $y \in Y=\{1,\dots,C\}$, with $C$ being the number of classes. We are interested in learning the parameters $\theta$ of a function $f_\theta:\mathcal{X}\longrightarrow \mathcal{Y}$ mapping images to probability vectors over the class labels, with $\mathcal{Y}$ being a probability simplex defined over $Y$. Note that, at test time, we receive images $X_t$ from a new dataset $\mathcal{D}_t$, with a different joint distribution, \ie $p_{\mathtt{x}\mathtt{y}}^\mathcal{D}\neq p_{\mathtt{x}\mathtt{y}}^{\mathcal{D}_t}$, with $\mathtt{x}$ and $\mathtt{y}$ being random variables in $\mathcal{X}$ and $Y$ respectively.  

We train our model $\theta$ based on two simple principles: 1) simulating the presence of multiple domains via a set of data augmentations, 2) enforcing that the output of the model is consistent across original and simulated domains. Formally, we define our overall learning objective function as:
\begin{equation}
\label{eq:general-framework}
\mathcal{L} = \sum_{(X,y)\in\mathcal{D}} \mathcal{L}_\text{CE}(X,\phi(X), y) + \lambda \mathcal{L}_\text{CON}(X,\phi(X),y), 
\end{equation}
where $\phi$ is a label-preserving augmentation function, $\mathcal{L}_\text{CON}$ is a consistency term between $X$ and its augmented version $\phi(X)$ given the semantic label $y$, and $\lambda$ is a hyperparameter balancing the two loss terms. $\mathcal{L}_\text{CE}$ is the cross-entropy loss:
 \begin{equation}
\mathcal{L}_{\text{CE}}(X,\hat{X},y) = -\log f^y_{\theta}(X) - \log f^y_{\theta}(\hat{X}),
\label{eq:loss_ce}
\end{equation}
where $f_\theta^y(X)$ is the probability of class $y$ for the input $X$ given by the function $f_\theta$.

The form of $\phi$ and $\mathcal{L}_\text{CON}$ influence the performance of the framework. In this work, we randomly sample $\phi$ from a larger set $\Phi$ composed of visual corruptions, \ie transformations that alter the content of the image while not modifying the location of the object of interest. These corruptions provide large visual variations while being simple and efficient w.r.t. other state-of-the-art alternatives. Since the locations of the objects are preserved, we can implement $\mathcal{L}_\text{CON}$ by 1) extracting the spatial regions that most contributed to the prediction and 2) enforcing the model to focus on the same regions, independent of the specific corruption of the input. As we will show experimentally, this supervision is more effective than enforcing consistency on 
model's predictions. 

\subsection{Visual Corruptions}
\label{sec:visco_modes}

Here we describe our set of transformations $\Phi$, merging the ImageNet-C with Fourier transform-based corruptions. 

\subsubsection{ImageNet-C Visual Corruptions}
\label{sec:base_corruptions}

ImageNet-C \cite{hendrycks2018benchmarking} is a well-known benchmark to evaluate the robustness of visual models under corruptions \cite{zhaoNIPS20maximum,hendrycks2020augmix,michaelis2019dragon}. It contains $19$ corruptions in total, with $5$ severity levels. We argue that corruptions can be used as an augmentation technique to train robust vision models. The corruptions in ImageNet-C are grouped into four categories, \ie Weather, Blur, Noise and Digital (see Figure \ref{fig:VisCo} for examples).

\textbf{Weather} simulates meteorological hurdles such as \textit{fog}, \textit{snow}, \textit{frost} and \textit{spatter} whereas \textbf{Blur} smooths the intensities of the image pixels using different functions, such as \textit{gaussian}, \textit{glass}, \textit{motion}, \textit{defocus} and \textit{zoom}. \textbf{Noise} perturbates the pixel values randomly, using different functions, \ie \textit{shot}, \textit{impulse}, \textit{Gaussian} and \textit{speckle} while \textbf{Digital} gathers diverse set of corruptions caused by either modifying the image resolution (\ie \textit{JPEG compression}, \textit{pixelation}, \textit{elastic}) or pixel intensity (\ie \textit{saturation}, \textit{brightness}, and \textit{contrast}).

\subsubsection{Fourier-based Visual Corruptions}
\label{sec:fourier_based_corruptions}

Early studies showed how the phase component of Fourier transform of images retains most of the semantic in a scene whereas amplitude focuses on textures \cite{piotrowski1982demonstration}. Recent works successfully used this property in domain adaptation \cite{yang2020phase,yang2020fda} and multi-source domain generalization \cite{xu2021fourier}. We thus incorporate three frequency-based corruption methods to our pool of transformations. In the following we use $\mathcal{F}(X)$ to denote the Fourier transform of an image $X$, with $\mathcal{F}^A(X)$ its amplitude and with $\mathcal{F}^P(X)$ its phase. 

\myparagraph{Phase Scaling}. 
Given a random scalar $\alpha \in (0, 1]$, this corruption uses $\alpha$ to scale the phase component, computing:
\begin{equation}
    \phi_\text{P-scaling}(X) = \mathcal{F}^{-1}([\mathcal{F}^A(X),\alpha \mathcal{F}^P(X)]), 
\end{equation}
where $\mathcal{F}^{-1}$ is the inverse Fourier transform. By scaling the phase, we are adding more visual artifacts that will occlude elements of the scene as $\alpha\rightarrow 0$ (Fig.~\ref{fig:VisCo}, first Fourier sample).

\myparagraph{Constant Amplitude}. 
This corruption replaces $\mathcal{F}^A$ with a constant $\beta \in (0,1]$, computing the corrupted image as:
\begin{equation}
    \phi_\text{constant-A}(X) = \mathcal{F}^{-1}([\beta, \mathcal{F}^P(X)]), 
\end{equation}
Since phase information is preserved, the resulting images are recognizable, but lose most color and texture information (Figure \ref{fig:VisCo}, second Fourier sample). 

\myparagraph{High pass filter}. 
This transformation corrupts the input image with a high pass filter via frequency windows. It filters out low frequency components by adjusting its diameter $d$ on the centered Fourier spectrum. Formally:
\begin{equation}
    \label{eq:frequency-filtering}
    \phi_\text{high-pass}(X) = \mathcal{F}^{-1}(H^d(\mathcal{F}(X)) \circ \mathcal{F}(X))),  
\end{equation}
where $H^d(F)$ a filtering mask where each spatial coordinate $(u,v)$ has value:
\begin{equation}
    \label{eq:high-pass}
     H_{u,v}^d(F) = \begin{cases} 1, & \text{if} \;\;\; F_{u,v}\geq d\\ 
     0, & \text{otherwise.}
    \end{cases}
\end{equation} 
This leads to a corrupted image where edges are highlighted and shapes are preserved (Figure \ref{fig:VisCo}, third Fourier sample).

\subsection{Attention Consistency} 
\label{sec:objective}

Visual corruptions provide powerful augmentations for single DG. However, 
we argue that a good single DG model should also look at the same image regions, no matter of their particular style. This will allow the model to find consistent visual cues across different versions of the same input, re-using these cues in unseen target domain. 
In this section, we describe how to use CAMs of original and corrupted images to define a consistency loss term for single DG. 

\myparagraph{CAM consistency.}
CAMs \cite{zhou2016learning} provide visual explanations to a given model's predictions by visualizing the spatial regions that most contributed to the output in a given feature map. Let us split $f_\theta$ in three components: $g:\mathcal{X}\rightarrow\mathcal{Z}$ mapping an image into the feature space $\mathcal{Z}\subset \mathbb{R}^{n\times s}$, an average pooling operation $P$ and a linear classifier $W \in \mathbb{R}^{n\times C}$ followed by softmax. In $\mathcal{Z}$,  $n$, denotes number of channels, and $s$ the spatial locations.  Following the formulation of \cite{sun2020fixing}, 
given an input $X$ and we define its set of CAMs as: 
\begin{equation}
\label{eq:cam}
    M = \sigma(W^\intercal g(X)),
\end{equation}
where $M \in \mathbb{R}^{C\times s}$ and $M_c\in \mathbb{R}^{s}$ denotes the CAM for class $c$, corresponding to the $c$-th row of $M$.  In Eq.~\eqref{eq:cam}, $\sigma$ is a softmax operation with temperature $T$ over the locations:
\begin{equation}
    \sigma(x)^c_i = \frac{exp(x^c_i / T)}{\sum_{j=1}^{s}exp(x^c_j / T))}.
\end{equation}

Given a label $y$ we compute our visual attention consistency loss using Jensen-Shannon divergence as:
\begin{equation}
    \mathcal{L}_{\text{CAM}}(M,\hat{M},y) = D_{JS}(M_y || \hat{M}_y),
\label{eq:loss_cam}
\end{equation}
where $\hat{M_y}$ is the CAM of the corrupted image $\hat{X}=\phi(X)$ for the class $y$. While Eq.~\eqref{eq:loss_cam} can be replaced by other objectives, such as MSE, we found the Jensen-Shannon divergence (JSD) to work better in practice. Moreover, this formulation allows to define more flexible objectives through the temperature $T$ of the softmax, since $T < 1$ leaves only the extreme points of attention whereas $T > 1$ smooths the CAM over the image.

\myparagraph{Negative CAM loss.} One problem with CAMs is that models tend to produce false activations, \ie attention maps localized in precise regions even when a class is not present in the input image  \cite{sun2020fixing}. Since our consistency loss heavily relies on the quality of the CAMs, we use the negative CAM loss \cite{sun2020fixing} to penalize attention maps for absent classes in the input (Figure \ref{fig:VisCo}, red box). The loss is defined as:
\begin{equation}
\mathcal{L}_{\text{NEG}}(M,C_k) = 
\sum_{c \in C_k} D_{KL}(U || M_c) + D_{KL}(U || \hat{M}_c), 
\label{eq:loss_neg}
\end{equation}
where $U$ is the uniform distribution over the spatial locations $s$, and $C_k$ is the set of top-$k$ negative classes in terms of their confidence scores for the clean image $X$. From \cref{eq:loss_cam,eq:loss_neg}, we can define our final objective as: 
\begin{equation}
\mathcal{L} = \mathcal{L}_{\text{CE}} + \lambda (\mathcal{L}_{\text{CAM}} + \mathcal{L}_{\text{NEG}}).
\label{eq:loss}
\end{equation}
We name our final model \fullnameBold\ 
(\textbf{\nickname}). 
 
\begin{algorithm}[t!]
\caption{Single DG with \nickname}
\label{alg:alg1}     
\begin{algorithmic}[1]
\REQUIRE Training set $\mathcal{D}$, parameters $\theta$, set of corruptions $\Phi$, prediction function $f$.
    \FORALL {$(X,y) \in \mathcal{D}$}
        \STATE {Randomly sample a corruption operation $\phi$ from $\Phi$} 
        \STATE {Apply the transformation to the input: $\hat{X} = \phi(X)$}
        \STATE Compute predictions $f_\theta(X)$, $f_\theta(\hat{X})$
        \STATE Compute CAMs $M$, $\hat{M}$ using Eq.\eqref{eq:cam}
        \STATE {Compute the loss: $\mathcal{L} = \mathcal{L}_{\text{CE}} + \lambda( \mathcal{L}_{\text{CAM}} + \mathcal{L}_{\text{NEG}})$}
        \STATE {Compute the gradient of $\theta$ w.r.t. $\mathcal{L}$}
        \STATE {Update $\theta$}
    \ENDFOR
\end{algorithmic}
\end{algorithm} 

\begin{table*}[!t]
\centering
	 \centering
		 \begin{tabular}{lc|ccccc} 
		 \toprule
		 & Photo & Art & Cartoon & Sketch & Avg. & Max. \\
		 \midrule
		 Baseline                                         & $98.52 \pm 0.4$ & $55.62 \pm 2.2$ & $18.56 \pm 2.6$ & $25.81 \pm 4.8$ & $33.33 \pm 2.4$ & $37.11$ \\
		 MixUp \cite{zhang2017mixup}                      & $97.32 \pm 0.7$ & $52.82 \pm 0.7$ & $16.97 \pm 4.4$ & $23.21 \pm 4.5$ & $31.00 \pm 1.7$ & $32.83$ \\
		 CutOut \cite{devries2017improved}                & $98.49 \pm 0.6$ & $59.84 \pm 1.3$ & $21.56 \pm 1.6$ & $28.83 \pm 3.3$ & $36.74 \pm 1.5$ & $39.24$ \\
		 CutMix \cite{yun2019cutmix}                      & $98.20 \pm 0.6$ & $59.63 \pm 1.8$ & $21.98 \pm 3.9$ & $24.94 \pm 4.7$ & $35.52 \pm 2.3$ & $38.92$ \\
		 ME-ADA \cite{zhaoNIPS20maximum}                  & $96.49 \pm 0.8$ & $55.61 \pm 0.9$ & $28.92 \pm 1.5$ & $24.63 \pm 4.3$ & $36.39 \pm 1.8$ & $39.08$ \\
		 RandAugment \cite{cubuk2020randaugment}          & $\mathbf{99.22} \pm 0.6$ & $\mathbf{67.81} \pm 0.9$ & $28.94 \pm 2.6$ & $36.96 \pm 4.7$ & $44.57 \pm 2.3$ & $48.79$ \\
		 AugMix \cite{hendrycks2020augmix}                & $98.44 \pm 0.3$ & $63.94 \pm 1.6$ & $27.72 \pm 1.4$ & $30.86 \pm 3.2$ & $40.84 \pm 1.4$ & $43.11$ \\
		 \midrule
         VC  (Ours)                                       & $\underline{98.75} \pm 0.6$ & $67.23 \pm 0.5$ & $\underline{30.26} \pm 2.1$ & $\underline{43.81} \pm 3.9$ & $\underline{47.10} \pm 1.7$ & $\underline{49.48}$ \\
         ACVC (Ours)                                      & $\mathbf{99.22} \pm 0.4$ & $\underline{67.80} \pm 0.9$ & $\mathbf{30.31} \pm 2.1$ & $\mathbf{46.42} \pm 6.7$ & $\mathbf{48.18} \pm 2.8$ & $\mathbf{54.67}$ \\
		 \bottomrule
		 \end{tabular}
	\vskip -0.2em
	 \caption{Comparing with the state of the art on PACS benchmark on single DG task using ResNet-18. The models are trained on Photo domain, and tested on Art, Cartoon and Sketch domains. We measure classification accuracy. Baseline: ResNet-18 trained with cross-entropy loss w/o any augmentations. Best numbers are bold, second best are underlined. VC = ACVC w/o attention consistency.}
	 \label{tab:main_pacs}
\end{table*} 
 
\subsection{Algorithm Overview}
\label{sec:alg}

We summarize the model training in Algorithm \ref{alg:alg1}. As the algorithm shows, we first sample a training image and its label (line 1) from $\mathcal{D}$. We then sample a random corruption from our set $\Phi$ (line 2) and we apply the transformation to the input image (line 3). For each sample in the batch we compute its prediction on both original and corrupted samples (line 4) and their relative CAMs (line 5). Finally, we compute our loss using Eq.~\eqref{eq:loss} (line 6), the gradient of the parameters w.r.t. the loss (line 8) and update the parameters (line 9). During training, we apply the corruptions without any additional augmentation techniques. At inference, no corruption is applied on the images of unseen domains. 

\section{Experiments}
\label{sec:experiments}

\myparagraph{Datasets and setup.} We evaluate our model on three challenging benchmarks for single DG: PACS~\cite{li2017deeper}, COCO~\cite{lin2014microsoft}, and DomainNet~\cite{peng2019moment} in increasing order of difficulty. 

\textbf{PACS} is a standard multi-source domain generalization benchmark \cite{li2017deeper}, with $9{,}991$ images belonging to $7$ different classes. We use it in the single DG setting due to the extreme domain-shift between its four domains, \ie Photo, Art painting, Cartoon and Sketch. Since in this work we are specifically interested in generalizing from natural images, we consider Photo as the source domain. 

For \textbf{COCO}, we propose a new benchmark consisting of $10$ shared classes between the original MS-COCO~\cite{lin2014microsoft} and DomainNet~\cite{peng2019moment}. We take MS-COCO as the training set, and test with the six domains in DomainNet: Real, Infograph, Painting, Clipart, Sketch, and Quickdraw. This setting is similar to that of \cite{zunino2020explainable}, but we only use training images where the target object covers at least $10\%$ of the pixels. Since this constraint may limit the number of images of some classes, we avoid class imbalance by setting $1{,}000$ as the upper bound on the number of samples per class, obtaining $7{,}783$ images in total. As in \cite{zunino2020explainable}, we test on this benchmark since the available segmentation masks allows us (and potentially future works) to explore how modeling the location of an object can improve the single DG performance. 

Finally, we include a large scale investigation using the full \textbf{DomainNet} dataset. It has $345$ object classes and contains $596{,}010$ images in total. We use the Real domain for training and validation, and the other five for testing. This dataset is extremely challenging due to the high-variability of the domains and the large number of classes. 

For all settings, we resize the RGB images to $224 \times 224$, and use the official train/val/test splits. We employ an ImageNet \cite{russakovsky2015imagenet} pretrained ResNet-18 \cite{he2016deep}, and use SGD optimizer with a learning rate of $4 \times 10^{-3}$, a batch size of $128$ and we train for $30$ epochs, dropping the learning rate by $0.1$ after $24$ epochs. These are the hyperparameters proposed by \cite{huang2020self} for multi-source domain generalization using PACS, and we keep these hyperparameters constant across our three benchmarks. In addition, for \nickname, we set $k=3$ empirically, and $\lambda = 0.06$ as in \cite{sun2020fixing}. For our Fourier-based corruptions, we define $5$ severity levels for $\alpha$, $\beta$, and $d$, as in ImageNet-C (see supplementary). During training, we randomly sample the severity of both ImageNet-C and Fourier-based operations uniformly from these $5$ levels. 

\myparagraph{Baselines and metrics.} 
We establish the single DG performance comparison using (1) a deep neural network trained using cross-entropy loss but without any data augmentation ({Baseline}), (2) advanced augmentation techniques, \ie MixUp \cite{zhang2017mixup}, CutOut \cite{devries2017improved}, CutMix \cite{yun2019cutmix}, (3) methods that randomly select augmentations from a large pool of transformations (where most corruption operations are omitted), \ie RandAugment \cite{cubuk2020randaugment}, AugMix \cite{hendrycks2020augmix}, and (4) the state-of-the-art adversarial data augmentation technique, \ie ME-ADA \cite{zhaoNIPS20maximum}. These methods do not have any reported results on our benchmarks, hence, we run our own experiments using the authors' implementations and suggested configurations if applicable. We provide mean accuracy and standard deviation measurements for multiple runs and the maximum achievable average domain generalization performance per method. The code will be released upon acceptance.

\begin{table*}[!t]
\centering
\setlength{\tabcolsep}{4pt}
    \begin{adjustbox}{width=1\textwidth}
	 \centering
		 \begin{tabular}{lc|cccccccc} 
		 \toprule
		 & COCO & Real & Painting & Infograph & Clipart & Sketch & Quickdraw & Avg. & Max. \\
		 \midrule
		 Baseline                                         & $80.44 \pm 0.7$ & $84.15 \pm 0.7$ & $78.55 \pm 0.5$ & $31.56 \pm 2.1$ & $62.90 \pm 3.4$ & $44.93 \pm 1.7$ & $12.56 \pm 2.4$ & $52.44 \pm 1.0$ & $54.46$ \\
		 MixUp \cite{zhang2017mixup}                      & $80.79 \pm 0.7$ & $78.61 \pm 1.0$ & $73.70 \pm 1.0$ & $23.96 \pm 1.1$ & $50.39 \pm 2.7$ & $38.82 \pm 1.7$ & $13.59 \pm 1.1$ & $46.51 \pm 0.8$ & $48.21$ \\
		 CutOut \cite{devries2017improved}                & $\underline{80.93} \pm 0.4$ & $84.17 \pm 0.4$ & $79.58 \pm 0.8$ & $32.45 \pm 1.7$ & $61.62 \pm 2.6$ & $39.73 \pm 3.2$ & $10.42 \pm 0.6$ & $51.33 \pm 1.1$ & $53.35$ \\
		 CutMix \cite{yun2019cutmix}                      & $80.13 \pm 0.6$ & $84.03 \pm 0.9$ & $78.72 \pm 0.7$ & $31.73 \pm 1.4$ & $64.08 \pm 2.8$ & $43.35 \pm 2.1$ & $12.22 \pm 0.9$ & $52.35 \pm 0.7$ & $53.88$ \\
		 ME-ADA \cite{zhaoNIPS20maximum}                  & $78.35 \pm 0.9$ & $82.28 \pm 1.0$ & $77.69 \pm 0.5$ & $28.58 \pm 1.4$ & $63.88 \pm 1.8$ & $45.29 \pm 1.7$ & $12.32 \pm 1.1$ & $51.67 \pm 0.8$ & $52.71$ \\
		 RandAugment \cite{cubuk2020randaugment}          & $80.51 \pm 0.6$ & $85.55 \pm 0.6$ & $\underline{81.67} \pm 0.4$ & $\underline{33.87} \pm 0.9$ & $67.96 \pm 2.8$ & $52.58 \pm 1.4$ & $14.57 \pm 1.3$ & $56.03 \pm 0.5$ & $56.75$ \\
		 AugMix \cite{hendrycks2020augmix}                & $80.50 \pm 0.6$ & $\underline{85.60} \pm 0.6$ & $80.19 \pm 0.7$ & $33.50 \pm 1.6$ & $71.37 \pm 0.8$ & $51.96 \pm 1.3$ & $17.96 \pm 2.3$ & $56.76 \pm 0.7$ & $57.79$ \\
		 \midrule
		 VC (Ours)                                        & $80.58 \pm 0.5$ & $\mathbf{85.86} \pm 0.7$ & $80.94 \pm 0.6$ & $32.50 \pm 1.0$ & $\underline{71.93} \pm 1.4$ & $\underline{61.98} \pm 1.8$ & $\underline{18.42} \pm 1.4$ & $\underline{58.61} \pm 0.7$ & $\underline{59.63}$ \\
		 ACVC (Ours)                                      & $\mathbf{81.80} \pm 0.6$ & $85.27 \pm 0.5$ & $\mathbf{82.37} \pm 0.6$ & $\mathbf{35.40} \pm 0.6$ & $\mathbf{73.04} \pm 0.8$ & $\mathbf{62.72} \pm 1.0$ & $\mathbf{21.25} \pm 0.9$ & $\mathbf{60.01} \pm 0.3$ & $\mathbf{60.23}$ \\
		 \bottomrule
		 \end{tabular}
	 \end{adjustbox}
	 \vskip -0.2em
	 \caption{Comparing with the state of the art on COCO benchmark on single DG task using ResNet-18. The models are trained on COCO dataset, and tested on DomainNet dataset. We measure classification accuracy. Baseline: ResNet-18 trained with cross-entropy loss only w/o any augmentations. Bold figures are the highest numbers, underlined are the second highest. VC = ACVC w/o attention consistency. }
	 \vspace{-10pt}
	 \label{tab:main_coco}
\end{table*}

\subsection{Comparison on PACS}
\label{sec:pacs}

Table \ref{tab:main_pacs} shows our evaluation on PACS, where there exists a large distribution shift between the source and target domains (\eg Photo to Sketch). Despite the large domain gap, our proposed methods surpass all competitors on this benchmark. Visual corruptions alone (\nicknameCorruptions) obtain superior performance on PACS, with an average accuracy of $47.10\% \pm 1.7$ across the different unseen domains in comparison to RandAugment with $44.57\% \pm 2.3$ showing its effectiveness for single DG. On average, due to our visual attention consistency loss, \nickname\ improves \nicknameCorruptions\ results by $1.08\%$, and on the best case, \nickname\ can go as high as $54.67\%$ average single DG performance. Moreover, the table shows how data augmentation methods that apply a single type of transformation do not provide enough input variations for training, with ME-ADA and CutOut achieving $36.39\%$ and $36.74\%$ respectively in average. However, applying multiple transformations per image may also hurt the performance, \eg RandAugment outperforms AugMix by $3.73 \%$ despite using a single transformation and no contrastive loss term. 

We see that, as the domain shift increases (Art $\rightarrow$ Sketch), standard deviations of all methods also increase, which is the reason behind the large gap between the average and maximum performance measurements. We believe this problem to be caused by the training set size, since PACS contains only $1{,}499$ Photo images. 

\begin{table*}[!t]
\centering
    \begin{adjustbox}{width=1\textwidth}
	 \centering
		 \begin{tabular}{lc|ccccccc} 
		 \toprule
		 & Real & Painting & Infograph & Clipart & Sketch & Quickdraw & Avg. & Max. \\
		 \midrule
		 Baseline                                         & $76.04 \pm 0.8$ & $38.05 \pm 0.8$ & $13.31 \pm 0.4$ & $37.89 \pm 1.2$ & $26.26 \pm 1.3$ & $3.36 \pm 0.2$ & $23.78 \pm 0.8$ & $24.34$ \\
		 MixUp \cite{zhang2017mixup}                      & $76.11 \pm 0.2$ & $38.60 \pm 0.1$ & $\mathbf{13.94} \pm 0.2$ & $38.02 \pm 0.8$ & $26.01 \pm 0.7$ & $3.71 \pm 0.3$ & $24.05 \pm 0.4$ & $24.45$ \\
		 CutOut \cite{devries2017improved}                & $\mathbf{76.96} \pm 0.8$ & $38.34 \pm 0.7$ & $13.69 \pm 0.4$ & $38.44 \pm 1.3$ & $26.24 \pm 0.8$ & $3.65 \pm 0.4$ & $24.07 \pm 0.7$ & $24.69$ \\
		 CutMix \cite{yun2019cutmix}                      & $75.79 \pm 0.7$ & $38.28 \pm 1.1$ & $13.45 \pm 0.5$ & $38.65 \pm 1.8$ & $26.85 \pm 1.5$ & $3.60 \pm 0.4$ & $24.17 \pm 1.1$ & $24.96$ \\
		 ME-ADA \cite{zhaoNIPS20maximum}                  & $74.27 \pm 0.1$ & $37.95 \pm 0.1$ & $13.12 \pm 0.0$ & $40.31 \pm 0.1$ & $26.79 \pm 0.1$ & $4.53 \pm 0.2$ & $24.54 \pm 0.0$ & $24.60$ \\
		 RandAugment \cite{cubuk2020randaugment}          & $\underline{76.70} \pm 0.4$ & $41.30 \pm 0.8$ & $13.57 \pm 0.3$ & $41.11 \pm 1.1$ & $30.40 \pm 1.0$ & $5.31 \pm 0.5$ & $26.34 \pm 0.7$ & $\underline{26.85}$ \\
		 AugMix \cite{hendrycks2020augmix}                & $76.27 \pm 0.1$ & $40.79 \pm 0.3$ & $\underline{13.89} \pm 0.1$ & $41.67 \pm 0.3$ & $29.80 \pm 0.2$ & $\underline{6.26} \pm 0.0$ & $26.48 \pm 0.2$ & $26.61$ \\
		 \midrule
		 VC (Ours)                                        & $75.91 \pm 0.3$ & $\mathbf{41.38} \pm 0.3$ & $13.58 \pm 0.3$ & $\underline{41.80} \pm 0.7$ & $\underline{30.58} \pm 0.5$ & $6.06 \pm 0.4$ & $\underline{26.68} \pm 0.2$ & $\mathbf{26.91}$ \\
		 ACVC (Ours)                                      & $76.16 \pm 0.5$ & $\underline{41.32} \pm 0.6$ & $12.89 \pm 0.6$ & $\mathbf{42.79} \pm 0.3$ & $\mathbf{30.86} \pm 0.5$ & $\mathbf{6.57} \pm 0.5$ & $\mathbf{26.89} \pm 0.0$ & $\mathbf{26.91}$ \\
		 \bottomrule
		 \end{tabular}
    \end{adjustbox}
	 \caption{Comparing with the state of the art on large-scale DomainNet benchmark on single DG task using ResNet-18. The models are trained on Real domain, and tested on Painting, Infograph, Clipart, Sketch and Quickdraw domains. We measure classification accuracy. Baseline: ResNet-18 trained with cross-entropy loss only w/o any augmentations. Bold figures are the highest numbers, underlined are the second highest. VC does not contain attention consistency. ACVC is our full model.}
	 \vspace{-10pt}
	 \label{tab:main_domainnet}
\end{table*}

\subsection{Comparison on COCO}
\label{sec:coco}

Table \ref{tab:main_coco} shows our results on the COCO benchmark. \nicknameCorruptions\ alone again outperforms the competitors with an average single DG accuracy of $58.61\% \pm 0.7$ where the best method in literature, AugMix achieves $56.76\% \pm 0.7$. Combined with visual attention consistency, \ie \nickname, the average performance reaches to $60.01\% \pm 0.3$, improving the \nicknameCorruptions\ accuracy by $1.4\%$. Contrary to PACS, we see that with enough training data ($7{,}783$ images from COCO), the standard deviation of \nickname's performance is relatively small. 

According to the results, corruptions help generalizing to distant domains such as Sketch and Quickdraw. Note that the common subset of COCO and DomainNet datasets includes classes such as \textit{bus}, \textit{car} and \textit{truck} which often have Quickdraw examples that are easily confused one with another. Therefore, any improvement on this domain tends to be limited for this particular benchmark. Nevertheless, \nickname\ can provide $3.29\%$ improvement over AugMix ($17.96\%$) on Quickdraw. Another interesting observation is that, in addition to the single DG performance rankings, the ranking between different methods change even between COCO dataset and Real domain of DomainNet. For instance, MixUp accuracy decreases $80.79\% \rightarrow 78.61\%$ where ME-ADA accuracy increases $78.35\% \rightarrow 82.28$. Note that, COCO dataset is designed to have multiple target classes in a given scene, whereas Real domain of DomainNet contains mostly centralized images w.r.t. the object of interest, thus performance may increase when testing on the latter.

\subsection{Comparison on DomainNet}
\label{sec:domainnet}

Table \ref{tab:main_domainnet} shows our results on large-cale DomainNet benchmark. This is the most challenging setting, due to the large domain-shift among domains (\eg Real to Infograph, Real to Quickdraw), and the large number of classes (345). Even in this benchmark, visual corruptions alone (\nicknameCorruptions) improve single DG performance, achieving $26.68\%$ accuracy with the best competitor (AugMix) achieving $26.48\%$. 

When visual attention consistency is used (\nickname), the avg. single DG accuracy reaches $26.89\%$. For Quickdraw images, methods with additional supervision signal tends to perform better, \ie AugMix and \nickname achieving $6.26\%$, and $6.57$ accuracy, respectively. Table \ref{tab:main_domainnet} validates once again the importance of simulating different visual variations by collecting a set of transformations: the gap between the best single (and adversarial) augmentation technique (ME-ADA) and \nicknameCorruptions\ is more than $2\%$ on average. 

Finally, this benchmark reveals that even though all methods perform relatively well on the source domain ($[74.27\%, 76.96\%]$), we still do not have robust vision models since their performance significantly drops as the domain shift increases, \eg as in Infograph and Quickdraw cases. The former shows a model's ability to filter out texts, charts and other irrelevant sources of information to focus on the object, and the performance of all methods drops to the range $[12.89\%, 13.94\%]$. The latter contains mostly primitive drawings to represent an object without color, texture or background, and the range of classification accuracy becomes $[3.36\%, 6.57\%]$.

\subsection{Ablation Study}
\label{sec:ablation}
{
\setlength{\tabcolsep}{5pt}
\renewcommand{\arraystretch}{1.1}
\begin{table}
	 \centering
		 \begin{tabular}{lcc} 
		 \toprule
		 & PACS & COCO \\
		 \midrule
		 Baseline                 				& $33.33 \pm 2.4$ & $52.44 \pm 1.0$ \\
		 \midrule 
         Weather                                & $40.36 \pm 2.3$ & $\underline{55.69} \pm 0.4$ \\
         Blur                                   & $36.83 \pm 1.3$ & $53.39 \pm 0.1$ \\
         Noise                                  & $35.53 \pm 1.9$ & $53.21 \pm 0.7$ \\
         Digital                                & $39.79 \pm 3.4$ & $55.12 \pm 0.7$ \\
		 Fourier                      		    & $34.15 \pm 1.5$ & $54.18 \pm 0.4$ \\
		 \midrule
		 ImageNet-C                             & $\underline{42.12} \pm 2.5$ & $55.52 \pm 0.9$ \\
         VC                                     & $\mathbf{47.10} \pm 1.7$ & $\mathbf{58.61} \pm 0.7$ \\
		 \bottomrule
		 \end{tabular}
	 \caption{Ablation study of different visual corruptions on PACS, and COCO. ImageNet-C contains Weather, Blur, Noise and Digital corruptions. VC contains all five, including Fourier category.}
	 \vspace{-10pt}
	 \label{tab:ablation_all}
\end{table}
}

\myparagraph{Corruptions.} 
Here we study the effect of (1) each visual corruption category; (2) ImageNet-C corruptions; and (3) our \nicknameCorruptions. As Table \ref{tab:ablation_all} shows, different corruptions work differently across domains. For instance, Noise and Fourier families perform well on COCO, but they offer limited improvement upon Baseline for PACS. For Blur, we see the opposite case: it improves PACS performance, but performs similar to Baseline on COCO benchmark. On the other hand, Weather and Digital categories perform close to full ImageNet-C category across all domains. Each category brings improvement over Baseline performance, however, randomly sampling transformations from all five (\nicknameCorruptions) consistently yields better performance than any individual family. We can also see that our additional Fourier-based visual corruptions bring a significant improvement over the original ImageNet-C family of transformations. In detail, \nicknameCorruptions, on average, improves ImageNet-C results by $4.98\%$ on PACS, and $2.81\%$ on COCO. This suggests that randomly combining multiple visual corruptions is the best choice when there is no prior knowledge on the target domains, merging the benefits of all families while diminishing the negative effects that single corruption categories may have in particular benchmarks. 

\begin{table}
\centering
	 \centering
		 \begin{tabular}{lcc} 
		 \toprule
		 & PACS & COCO \\
		 \midrule
         VC                                                                     & $47.10 \pm 1.7$ & $58.61 \pm 0.7$ \\
         \hspace{3pt} + $\mathcal{L}_{\text{JSD}}$                              & $\underline{47.39} \pm 2.6$ & $57.83 \pm 0.6$ \\
         \hspace{3pt} + $\mathcal{L}_{\text{MSE}}$                              & $42.91 \pm 1.8$ & $58.02 \pm 0.4$ \\
         \hspace{3pt} + $\mathcal{L}_{\text{NEG}}$                              & $43.00 \pm 0.9$ & $\underline{59.68} \pm 0.7$ \\
         \hspace{3pt} + $\mathcal{L}_{\text{CAM}}$                              & $46.55 \pm 2.7$ & $\underline{59.67} \pm 0.6$ \\
         \hspace{3pt} + segm. masks + $\mathcal{L}_{\text{NEG}}$   & N/A & $58.65 \pm 0.5$ \\
         \hspace{3pt} + $\mathcal{L}_{\text{CAM}} + \mathcal{L}_{\text{NEG}}$  (ACVC) & $\mathbf{48.18} \pm 2.8$ & $\mathbf{60.01} \pm 0.3$ \\
		 \bottomrule
		 \end{tabular}
	 \caption{Ablation study of the different loss terms reported on PACS and COCO benchmarks.}
	 \vspace{-10pt}
	 \label{tab:ablation_loss}
\end{table}

\myparagraph{Consistency loss.}
Here we study the effects of different consistency loss terms on the single DG performance when applied on top of \nicknameCorruptions. Table \ref{tab:ablation_loss} shows that not all consistency losses bring the same improvements over \nicknameCorruptions. For instance, JSD loss on model {predictions} for the original and augmented versions (as in \cite{hendrycks2020augmix}), slightly improves PACS results ($+0.9\%$), but degrades the performance on COCO ($-0.8\%$). When we apply our attention consistency loss $\mathcal{L}_{\text{CAM}}$, performance is significantly better ($+3.64\%$ on PACS, $+1.65\%$ on COCO) than simple MSE loss between CAMs. Nevertheless, $\mathcal{L}_{\text{CAM}}$ alone is still not robust, as it degrades the performance of \nicknameCorruptions\ by $0.55\%$ on PACS, but improves it by $1.08\%$ on COCO. This is also the case for improving the CAMs using only $\mathcal{L}_{\text{NEG}}$ without any consistency loss (\ie $-4.1\%$ on PACS but $+1.07\%$ on COCO w.r.t. \nicknameCorruptions). However, when we combine both terms, we achieve consistent improvement in both benchmarks, \ie $+1.08\%$ on PACS and $+1.4\%$ on COCO w.r.t. \nicknameCorruptions. Notably, the benefits of using an additional loss terms over \nicknameCorruptions\ do not generalize across PACS and COCO, except for \nickname. 

\begin{figure}
\centering
\includegraphics[scale=0.5]{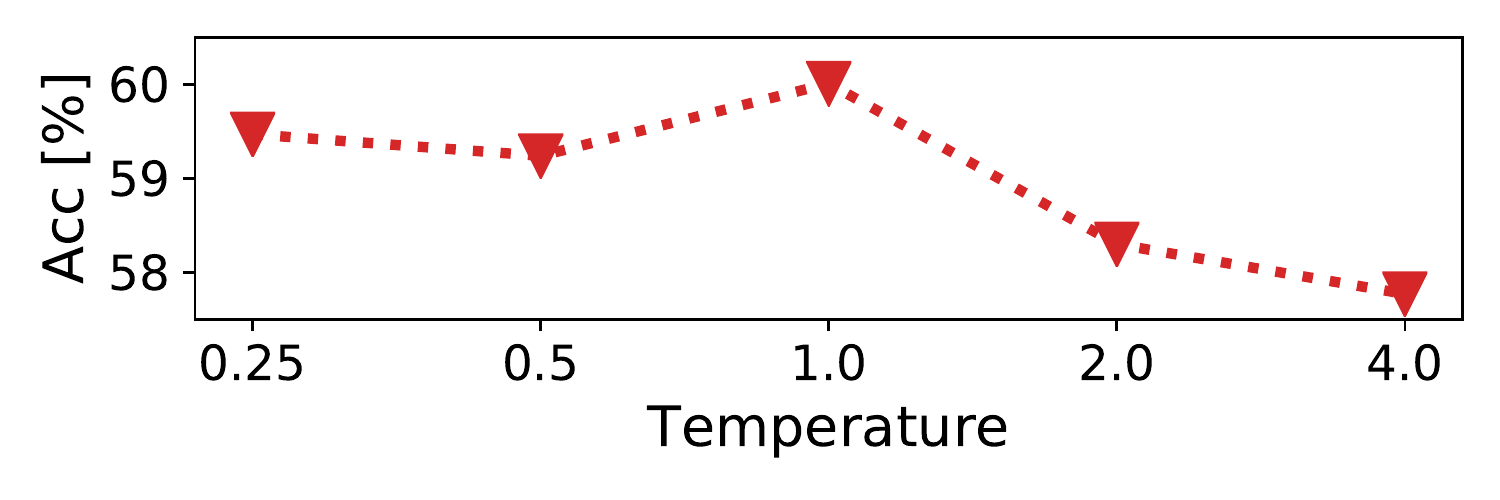}
\vspace{-13pt}
\caption{ACVC results on COCO benchmark for different $T$.}
\vspace{-13pt}
\label{fig:ablation_T}
\end{figure}

Finally, we analyze the effect of replacing our consistency loss on CAMs by imposing as fixed target in Eq.\eqref{eq:loss_cam} the normalized segmentation mask of the image provided by COCO dataset. Results show that $\mathcal{L}_{\text{CAM}}$ does not benefit from having a static reference point to optimize towards. On the contrary, \nickname\ is able to achieve, on average, $1.36\%$ higher single DG performance than using segmentation masks. We ascribe this behaviour to the nature of the softmax that spreads the intensity of the focus over the whole extent of the object and penalizes peaked values of the attention maps, even when they fall inside the object. This can also be seen on Figure \ref{fig:ablation_T}, where the avg. single DG performance is relatively better for $T \leq 1$, which shows how imposing consistency on peaks of the attention maps is more beneficial for single DG than smoothing the attention over larger spatial regions.

\subsection{Qualitative Results}
\label{sec:qualitative}

\begin{figure}
\centering
\includegraphics[scale=0.67]{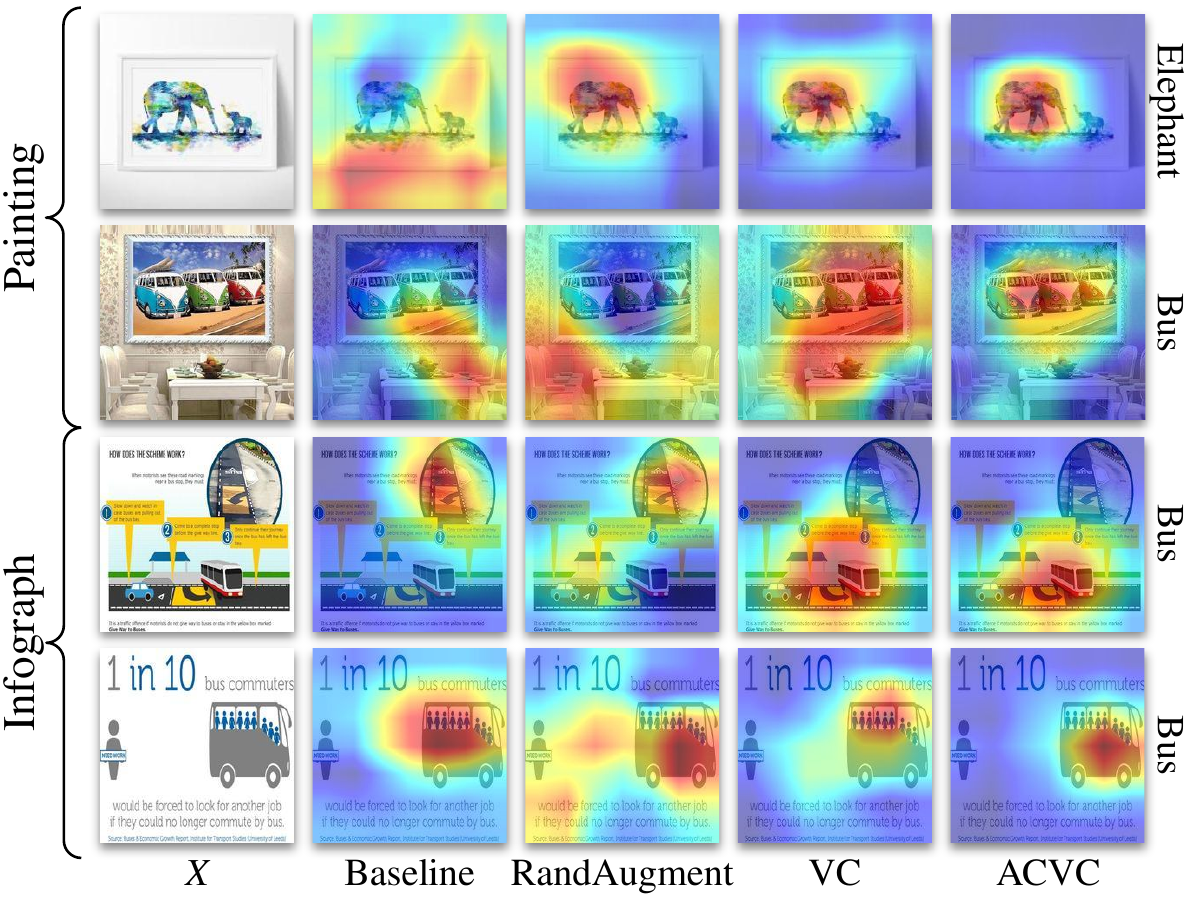}
\vspace{-5pt}
\caption{Class activation maps of (1) the baseline model, (2) two different sets of data augmentation techniques, \ie RandAugment and the proposed VC models, (3) attention consistency guided VC, \ie ACVC. Our ACVC approach obtains more fine-grained attention maps on unseen domains.}
\vspace{-10pt}
\label{fig:VisCo_CAM}
\end{figure}

In this section, we show CAMs for four different approaches, (1) the baseline model, (2) RandAugment and our VC as powerful pure data augmentation techniques, and (3) our final ACVC method. In Figure \ref{fig:VisCo_CAM}, we see that ACVC can recognize and focus on the relevant objects in unseen domains. In detail, the top two rows show paintings where ACVC is able to focus on the correct objects even in frames within a crowded scene. The last two rows show images from the challenging Infograph domain which contains charts, texts and symbols in addition to the target objects. Nevertheless, ACVC can still recognize the bus in both images.

\section{Conclusion}
\label{sec:conclusions}

In this work, we addressed the problem of single source domain-generalization (single DG) where the goal is to classify images of arbitrary unseen distributions, given a single domain at training time. Similar to previous works, we address the problem by synthesizing multiple training domains. However, unlike previous approaches, we propose to generate new domains by applying randomly sampled visual corruptions on the training data. Specifically, we consider a set of transformations that corrupt the original content in twenty-two different ways belonging to five categories of transformations (\ie  Weather, Blur, Noise, Digital, and Fourier). Since these transformations keep the object locations intact, we propose a visual attention consistency loss between the model's class activation maps for the original and corrupted versions of an input image. This loss ensures that the model focuses on the same image regions, disregarding the particular style of the input. Experiments show that our method, \nickname, consistently outperforms the state of the art in PACS, COCO and DomainNet benchmarks.

\myparagraph{Broader societal impact.}
Our method focuses on scenario where generalizing to unseen data distributions is crucial. As a consequence, \nickname\ can be applied in all scenarios involving robustness to different environmental conditions (\eg illumination, weather) as well as recognition across different visual modalities (\eg photo, cartoon, sketch). The ability to generalize to unseen domains without collecting additional unlabeled (as in domain adaptation \cite{csurka2017comprehensive}) or labeled (as in domain generalization \cite{li2017deeper}) data from different distributions could bring a positive impact on scenarios with privacy constraints (\eg federated learning \cite{li2020federated}), since it reduces the need of collecting data for specializing the recognition model to single users.  We want to highlight that the data used for experiments (PACS, COCO, and DomainNet) are all public datasets and do not contain any private information or disclose any identifiable personal information.

\myparagraph{Limitations.}
One limitation of our work is that we explicitly focus on generalizing from natural images, containing rich visual information. In this context, removing information through corruptions is beneficial for single DG performance. However, our approach may not be suitable for source domains where the input already presents limited information, such as sketches. In these cases we may need to replace our pool of corruptions with tailored augmentation techniques.

\myparagraph{Acknowledgements} 
This work has been partially funded by the ERC (853489 - DEXIM) and by the DFG (2064/1 – Project number 390727645).

{\small
\bibliographystyle{ieee_fullname}
\bibliography{egbib}
}

\title{Attention Consistency on Visual Corruptions \\ for Single-Source Domain Generalization\\ (Supplementary  Material)}

\author{Ilke Cugu$^1$, Massimiliano Mancini$^1$, Yanbei Chen$^1$, Zeynep Akata$^{1,2}$\\
$^1$University of T\"{u}bingen, $^2$MPI for Intelligent Systems\\
{\tt\small \{ilke.cugu, massimiliano.mancini, yanbei.chen, zeynep.akata\}@uni-tuebingen.de}
}
\maketitle


\setcounter{page}{1}
\setcounter{figure}{0}
\setcounter{section}{0}
\setcounter{table}{0}
\setcounter{algorithm}{0}
\renewcommand\thesection{\Alph{section}}
\renewcommand\thefigure{\Alph{figure}}
\renewcommand\thetable{\Alph{table}}
\renewcommand\thealgorithm{\Alph{algorithm}}

Here, we present more details on our experiments. We first provide detailed information on the hardware used for the experiments and the licenses of the datasets in Section \ref{sec:training_details}. Then, in Section \ref{sec:fourier_details}, we discuss the severity levels determined for our Fourier-based visual corruptions.

\section{Training Details}
\label{sec:training_details}

\myparagraph{Training GPUs.} All experiments are run by using $4 \times$ NVIDIA Quadro RTX 6000s, and $1 \times$ NVIDIA V100.

\begin{table}[!h]
    \begin{adjustbox}{width=0.46\textwidth}
	 \centering
		 \begin{tabular}{ll} 
		 \toprule
		 Dataset & Licence \\
		 \midrule
		 PACS               & Not available \\
		 COCO               & Creative Commons Attribution 4.0 License \\
		 DomainNet          & Custom: Non-commercial Research and Educational Purposes \\
		 \bottomrule
		 \end{tabular}
    \end{adjustbox}
	 \caption{The datasets employed in the paper and their licences.}
	 \label{tab:licence}
\end{table}

\myparagraph{Licences of the datasets.} In Table \ref{tab:licence}, we provide license information of the datasets we use in our experiments. Note that, for PACS, we did not find any attached license, but the dataset is publicly available\footnote{The dataset can be downloaded from the scripts in \url{https://github.com/liyiying/Feature_Critic}.}.

\section{Fourier-based Visual Corruptions}
\label{sec:fourier_details}

In this work, we propose to use three additional visual corruptions based on post-Fourier transform components along with ImageNet-C operations. ImageNet-C comes with $5$ severity levels for each operation. For compatibility, we also defined $5$ severity levels for each Fourier-based transformation. For all Fourier-based corruptions, we first set the highest severity level through visual inspection, ensuring that the images are highly corrupted but the objects are still easily recognizable to a human observer. Then, the intermediate levels are determined by dividing the interval between the clean image and the highest severity level into five equal parts. In the following we provide a summary of the transformations and their respective severity levels. We use $\mathcal{F}(X)$ to denote the Fourier transform of an image $X$, with $\mathcal{F}^A(X)$ its amplitude and with $\mathcal{F}^P(X)$ its phase.

\begin{figure}[!t]
\centering
\includegraphics[scale=0.65]{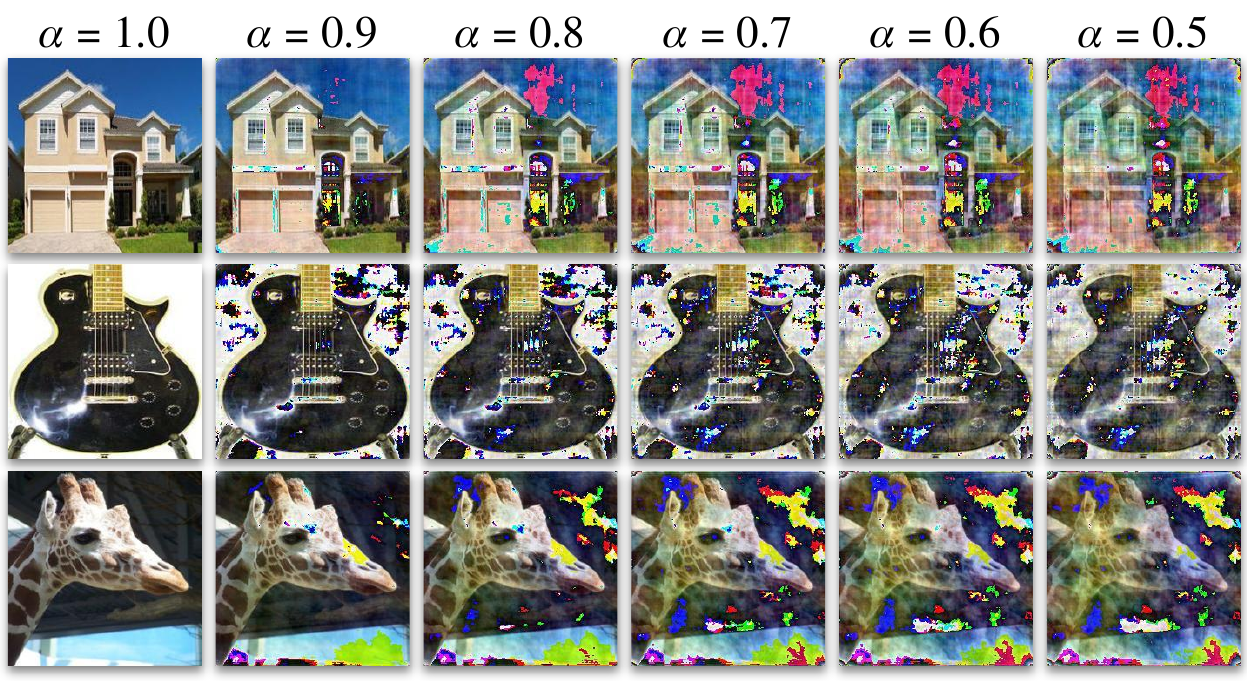}
\caption{Sample images of \textbf{phase scaling} corruption for $5$ severity levels. The intensity of corruption increases from left $\rightarrow$ right.}
\label{fig:phase_scaling}
\end{figure}

\myparagraph{Phase Scaling}.
Given a random scalar $\alpha \in (0, 1]$, this corruption uses $\alpha$ to scale the phase component, computing:
\begin{equation}
    \phi_\text{P-scaling}(X) = \mathcal{F}^{-1}([\mathcal{F}^A(X),\alpha \mathcal{F}^P(X)]), 
\end{equation}
where $\mathcal{F}^{-1}$ is the inverse Fourier transform, and the phase is computed by:
\begin{equation}
    \mathcal{F}_{u, v}^P(X) = \arctan \left( \frac{R(\mathcal{F}_{u, v}(X))}{I(\mathcal{F}_{u, v}(X))} \right)
\end{equation}
where $R$ is the real part, and $I$ is the imaginary part of $\mathcal{F}(X)$. In this work, we set minimum value of $\alpha$ as $0.5$, and the severity levels as $\{0.9, 0.8, 0.7, 0.6, 0.5\}$. In Figure \ref{fig:phase_scaling}, we show the difference between the severity levels.

\begin{figure}[!t]
\centering
\includegraphics[scale=0.65]{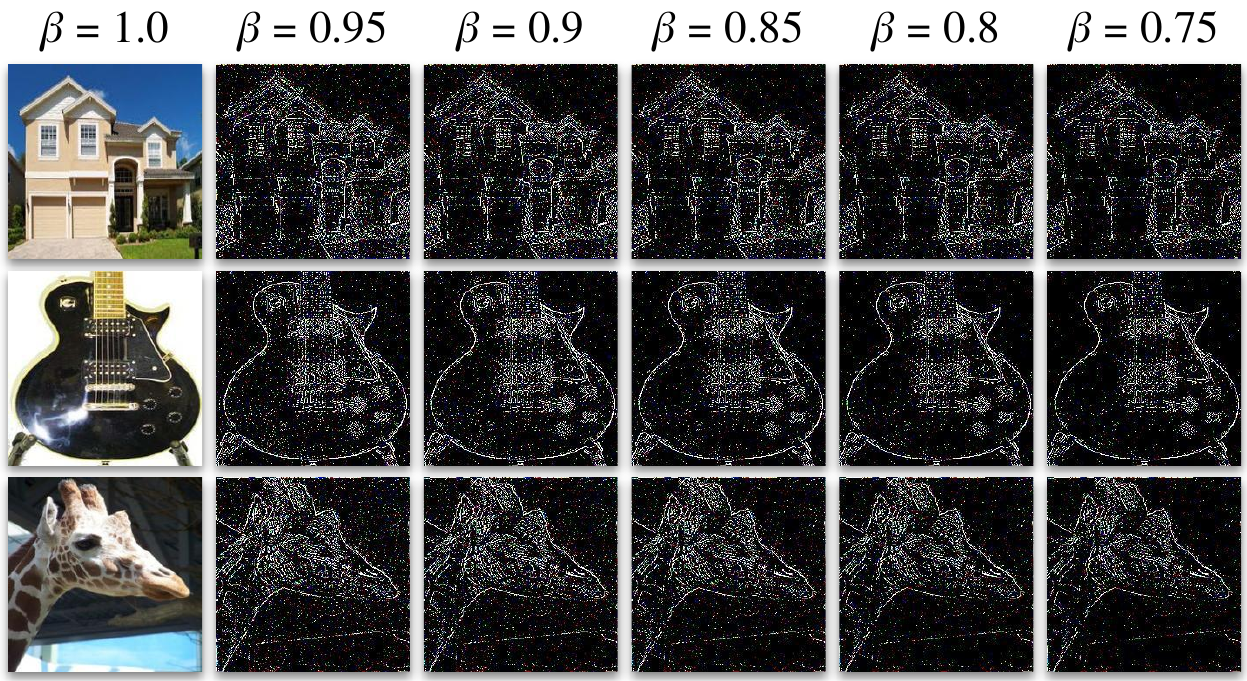}
\caption{Sample images of \textbf{constant amplitude} corruption for $5$ severity levels. The intensity of corruption increases from left $\rightarrow$ right.}
\label{fig:constant_amplitude}
\end{figure}

\myparagraph{Constant Amplitude}.
This corruption replaces $\mathcal{F}^A$ with a constant $\beta \in (0,1]$, computing the corrupted image as:
\begin{equation}
    \phi_\text{constant-A}(X) = \mathcal{F}^{-1}([\beta, \mathcal{F}^P(X)]), 
\end{equation}
where the amplitude is computed by:
\begin{equation}
    \mathcal{F}_{u, v}^A(X) = \sqrt{R^2(\mathcal{F}_{u, v}(X)) + I^2(\mathcal{F}_{u, v}(X))}
\end{equation}
In our experiments, $\beta$ can be $\{0.95, 0.9, 0.85, 0.8, 0.75\}$, with $0.75$ being the maximum corruption level.
In Figure \ref{fig:constant_amplitude}, we show the visual effects of these values. 

\begin{figure}[t]
\centering
\includegraphics[scale=0.65]{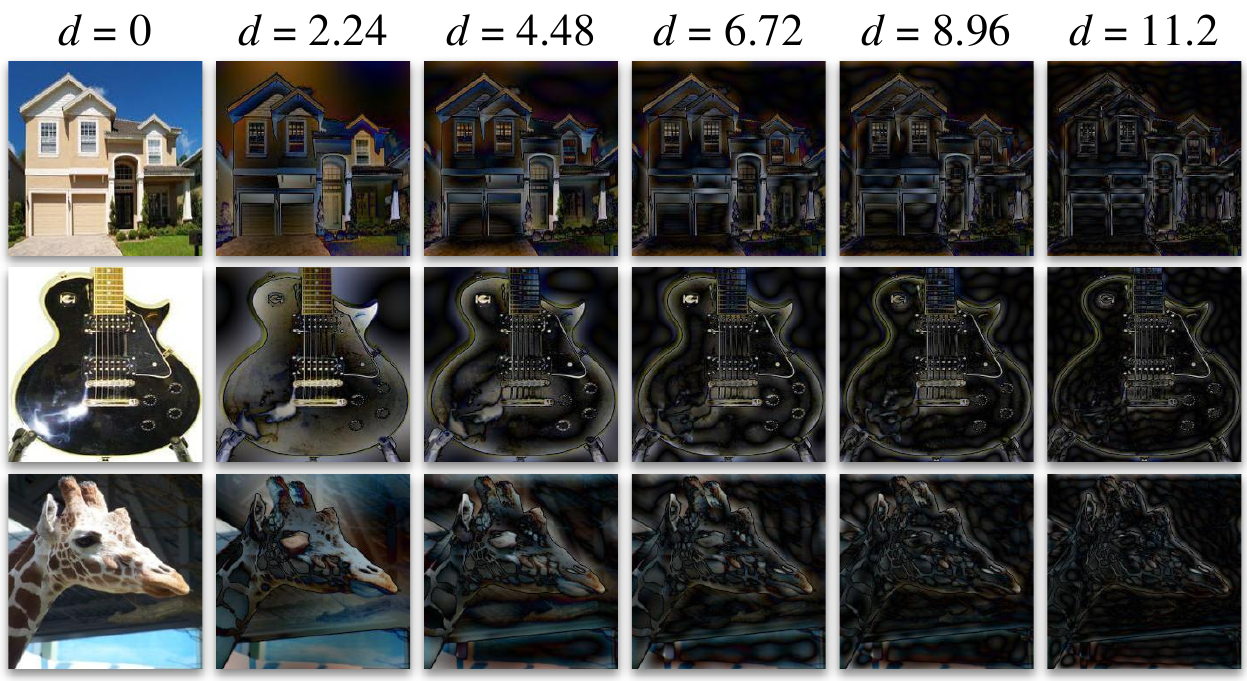}
\caption{Sample images of \textbf{high pass filter} corruption for $5$ severity levels. The intensity of corruption increases from left $\rightarrow$ right.}
\label{fig:high_pass_filter}
\end{figure}

\myparagraph{High pass Filter}.
This transformation corrupts the input image with a high pass filter via frequency windows. It filters out low frequency components by adjusting its diameter $d$ on the centered Fourier spectrum. Formally:
\begin{equation}
    \label{eq:frequency-filtering}
    \phi_\text{high-pass}(X) = \mathcal{F}^{-1}(H^d(\mathcal{F}(X)) \circ \mathcal{F}(X))),  
\end{equation}
where $H^d(F)$ a filtering mask where each spatial coordinate $(u,v)$ has value:
\begin{equation}
    \label{eq:high-pass}
     H_{u,v}^d(F) = \begin{cases} 1, & \text{if} \;\;\; F_{u,v}\geq d\\ 
     0, & \text{otherwise.}
    \end{cases}
\end{equation} 
where $d$ is proportional to the dimensions of the input images. Since we use $224 \times 224$ images, the respective values of $d$ are in the set $224 \times \{0.01, 0.02, 0.03, 0.04, 0.05\}$ where $224 \times 0.05 = 11.2$ is the maximum corruption level. In Figure \ref{fig:high_pass_filter}, we show the difference between the severity levels.

\end{document}